\pdfoutput = 1
\documentclass[sigconf]{acmart}
\usepackage{amsmath,graphicx}
\usepackage{algorithm}
\usepackage{algorithmic}

\usepackage{subfigure}

\usepackage{booktabs}  
\usepackage{threeparttable}  
\usepackage{multirow}
\usepackage{bm}
\usepackage{balance}

\AtBeginDocument{%
  \providecommand\BibTeX{{%
    \normalfont B\kern-0.5em{\scshape i\kern-0.25em b}\kern-0.8em\TeX}}}

\setcopyright{acmcopyright}
\copyrightyear{2021}
\acmYear{2021}
\acmDOI{10.1145/3459637.3482154}

\acmConference[CIKM '21] {Proceedings of the 30th ACM International Conference on Information and Knowledge Management}{November 1--5, 2021}{Virtual Event, QLD, Australia}
\acmBooktitle{Proceedings of the 30th ACM International Conference on Information and Knowledge Management (CIKM '21), November 1--5, 2021, Virtual Event, QLD, Australia}
\acmPrice{15.00}
\acmISBN{978-1-4503-8446-9/21/11}

\begin{document}
\fancyhead{}
\title{New Tight Relaxations of Rank Minimization for Multi-Task Learning}
%





\author{Wei Chang}
\affiliation{%
  \institution{Northwestern Polytechnical University}
  \city{Xi’an}
  \state{Shaanxi}
  \country{China}
}
\email{hsomewei@gmail.com}

\author{Feiping Nie}
\authornote{F. Nie is the corresponding author. This work was supported in part by the Innovation Foundation for Doctor Dissertation of Northwestern Polytechnical University under Grant CX2021088 and the Fundamental Research Funds for the Central Universities under Grant G2019KY0501.}
\affiliation{%
  \institution{Northwestern Polytechnical University}
  \city{Xi’an}
  \state{Shaanxi}
  \country{China}
}
\email{feipingnie@gmail.com}

\author{Rong Wang}
\affiliation{%
  \institution{Northwestern Polytechnical University}
  \city{Xi’an}
  \state{Shaanxi}
  \country{China}
}
\email{wangrong07@tsinghua.org.cn}

\author{Xuelong Li}
\affiliation{%
  \institution{Northwestern Polytechnical University}
  \city{Xi’an}
  \state{Shaanxi}
  \country{China}
}
\email{li@nwpu.edu.cn}



\begin{abstract}
Multi-task learning has been observed by many researchers, which supposes that different tasks can share a low-rank common yet latent subspace. It means learning multiple tasks jointly is better than learning them independently. In this paper, we propose two novel multi-task learning formulations based on two regularization terms, which can learn the optimal shared latent subspace by minimizing the exactly $k$ minimal singular values. The proposed regularization terms are the more tight approximations of rank minimization than trace norm. But it's an NP-hard problem to solve the exact rank minimization problem. Therefore, we design a novel re-weighted based iterative strategy to solve our models, which can tactically handle the exact rank minimization problem by setting a large penalizing parameter. Experimental results on benchmark datasets demonstrate that our methods can correctly recover the low-rank structure shared across tasks, and outperform related multi-task learning methods.

\end{abstract}





\maketitle

\section{Introduction}
\label{sec:intro}
Multi-task learning (MTL) \cite{caruana1997multitask, zhang2018overview} is an emerging machine learning research topic, that has been popularly studied in recent years and applied to many scientific applications, such as computer vision \cite{kang2011learning, wang2011sparse}, medical image analysis \cite{wang2012high, chen2019multi}, web system \cite{chapelle2010multi, Yang2021} and natural language processing \cite{ando2005framework, worsham2020multi}. For traditional supervised learning, each task is often learned independently, which ignores the correlations between tasks. However, in the last decades, it has been observed by many researchers \cite{obozinski2006multi, kim2010tree}, that if the correlated tasks for different purposes are learned jointly, one would benefit from the others under the common or shared information and representations.

For multi-task learning, the most important challenge is how to discover the task correlations such that most tasks can benefit from the joint learning. One common way to define the task correlations is to assume that related tasks share a common yet latent low-rank feature subspace \cite{argyriou2008convex,ando2005framework}. $\ell_{2,1}$-norm \cite{liu2009multi} and $\ell_1$-norm \cite{zhang2017learning} based methods were proposed, that utilize the sparse constraints to extract the optimal shared feature subspace. However, the sparse structure cannot represent the inherent structure of the shared subspace completely. Due to the property of sparse constraints, the inherent structure may be destroyed in the optimization process. Therefore, the low-rank constraint is more capable of capturing the inherent structure of the shared feature subspace.

To obtain the low-rank structure, one obvious way is to solve the rank minimization objective. But, it's an NP-hard problem to directly optimize the rank minimization problem. Hence, the trace norm \cite{srebro2005rank, argyriou2008convex} is utilized as a convex relaxation of rank function for learning the common subspace of multiple tasks. However, the trace norm is not a tight approximation of rank function. For example, if the largest singular values of a matrix change significantly, the trace norm will also change significantly based on its definition, but the rank remains unchanged. Hence, it may make the trace norm based methods unable to capture the intrinsic shared task structures efficiently in multi-task learning.

To address the mentioned problem, we proposed two novel regularizations based models to approximate the rank minimization problem. For our models, if the minimal singular values are suppressed to zeros, the rank would also be reduced. Compared to the trace norm, the new regularizations are the more tight approximations for rank constraint, which make our algorithms have the better ability of discovering the low-rank feature subspace. Besides, an iterative optimization algorithm based re-weighted method is proposed to solve our models, which can tactically avoid the NP-hard problem like the rank minimization based models. Experimental results on synthetic and real-world datasets show that our models consistently outperform the existing superior methods.

\section{Related Work}
\label{revisit}
In this section, we revisit two classical MTL approaches related to our models. Suppose there are $T$ tasks, the $t$-th task has $n_t$ training data points $X_t = [x_1^t, x_2^t, \dots, x_{n_t}^t] \in R^{d \times n_t}$ and the corresponding label matrix $Y_t \in R^{c_t \times n_t}$ is given. To capture the low-rank structure of the shared task subspace, a general multi-task learning model can be formulated as 
\begin{equation}
\label{rankbased}
\min \limits_{W = [W_1, \dots,W_T]} \sum \limits_{t=1}^T f(W_t^{'}X_t, Y_t)+\gamma \, rank(W),
\end{equation}
where $W_t \in R^{d \times c_t}$ is the projection matrix to be learned, $W\in R^{d \times c}$ and $c = \sum_t c_t$. The first term is the sum loss of the $T$ learning tasks and the second term is the rank constraint on the shared projection matrix $W$. The transpose of matrix $W_t$ is defined by the notation $W_t^{'}$ in this paper.

Problem \eqref{rankbased} is an NP-hard problem due to the rank function. To avoid this issue, $\ell_{2,1}$-norm \cite{liu2009multi} is utilized to instead of the rank function in problem \eqref{rankbased}. Under the $\ell_{2,1}$ norm $\|W\|_{2,1}$, the projection matrices $W_1,\dots,W_T$ have the same row sparsity due to the shared matrix $W$. In addition, the reference \cite{recht2010guaranteed} points out that trace norm $\|W\|_*$ is the best convex envelope of $rank(W)$. Hence, the trace norm is introduced into the low-rank based multi-task learning approach \cite{argyriou2008convex} instead of $rank(W)$ to pursue the latent structure of the whole projection matrix $W$.

\section{Proposed Formulation}
\label{model}
Although the trace norm performs well in the approximation of rank minimization problem, it's not the tight approximation for the rank function. Here is an example. Suppose $\sigma_i(W)$ is the $i$-th smallest singular value of $W$. For a low rank matrix $W$, the smallest singular values in front should be zeros. Concretely, if the rank of $W \in R^{d\times c}$ is $r$, the $k$ smallest singular values should be zeros, where $k=m-r, m = min(d,c)$. Note that $\|W\|_*=\sum \nolimits_{i=1}^m \sigma_i(W)$, if the largest $r$ singular values of a matrix $W$ with rank $r$ are changed significantly, then $\|W\|_*$ will be changed significantly. However, the rank of $W$ is not changed. Thus there is a big gap between the function $\|W\|_*$ and $rank(W)$. In order to reduce this gap, we propose two new regularizations as follows
\begin{equation}
\label{reg}
\sum \nolimits_{i=1}^k \sigma_i^2(W), \quad \sum \nolimits_{i=1}^k \sigma_i(W).
\end{equation}
Here, the singular values $\sigma_i(W), i = 1,2,\dots,m$, of the matrix $W$ are ordered from the small to large.


For these two regularizations, we focus on the $k$-smallest singular values of $W$ and ignore the largest singular values, which is closer to the rank function than trace norm. Although they are both close to the rank function, they are not the same in essence. These two regularizations are more general. If $k=m$, the first regularization in Eq. \eqref{reg} is Frobenius norm and the second becomes the trace norm, respectively.

Based on these two presented regularization terms, we further propose to solve the following problems for MTL
\begin{equation}
\label{MTL-1}
\min \limits_{W = [W_1,...,W_T]} \sum \limits_{t=1}^T f(W_t^{'}X_t,Y_t)+\gamma \sum \limits_{i=1}^k \sigma_i^2(W).
\end{equation}

\begin{equation}
\label{MTL-2}
\min \limits_{W = [W_1,...,W_T]} \sum \limits_{t=1}^T f(W_t^{'}X_t,Y_t)+\gamma \sum \limits_{i=1}^k \sigma_i(W).
\end{equation}
We can see that when $\gamma$ is large enough, then the $k$ smallest singular values of the optimal solution $W$ to problem \eqref{MTL-1} or problem \eqref{MTL-2} will be zeros since all the singular values of a matrix are non-negative. That is to say, when $\gamma$ is large enough, it's equivalent to constraint the rank of $W$ to be $r=m-k$ for problem \eqref{MTL-1} and \eqref{MTL-2}.

It seems difficult to directly solve the proposed models due to the regularizations on the $k$ smallest singular values, which are NP-hard problems. Hence, we proposed two novel optimization algorithms based re-weighted method to solve our models, which tactically avoid the difficulty of solving the original problem \eqref{MTL-1} and \eqref{MTL-2}. It's interesting to see that the proposed algorithms are very efficient and easy to implement.

\section{Optimization Algorithms}
\subsection{Optimization for Problem \eqref{MTL-1}}
\label{opt1}
Based on Ky Fan's theorem \cite{fan1949theorem}, we have
\begin{equation}
\label{Kanfan}
\sum \limits_{i=1}^k \sigma_i^2(W) = \min \limits_{F \in R^{d \times k}, F^{'}F=I} Tr(F^{'}WW^{'}F).
\end{equation}
Therefore, the problem \eqref{MTL-1} can be rewritten as
\begin{equation}
\label{MTL-1kanfan}
\min \limits_{\substack{W = [W_1,\dots,W_T], \\ F \in R^{d\times k}, F^{'}F=I}  } \sum \limits_{t=1}^T f(W_t^{'}X_t,Y_t)+\gamma Tr(F^{'}WW^{'}F).
\end{equation}

Compared with the original problem \eqref{MTL-1}, this problem is much easier to solve. We can apply the alternative optimization approach to solve this problem. 

When $W$ is fixed, the problem \eqref{MTL-1kanfan} becomes
\begin{equation}
\label{MTL-1F}
\min \limits_{F\in R^{d \times k}, F^{'}F=I} Tr(F^{'}WW^{'}F).
\end{equation}
It's easy to see that the optimal solution $F$ to problem \eqref{MTL-1F} is formed by the $k$ eigenvectors of $WW^{'}$ corresponding to the $k$ smallest eigenvalues.

When $F$ is fixed, problem \eqref{MTL-1kanfan} becomes 
\begin{equation}
\label{MTL-1W}
\min \limits_{W = [W_1,\dots,W_T]} \sum \limits_{t=1}^T f(W_t^{'}X_t,Y_t)+\gamma \sum \limits_{t=1}^T Tr(W_t^{'}FF^{'}W_t).
\end{equation}

In this paper, we focus on solving the regression problem. So the least square loss function $f(W_t^{'}X_t,Y_t)$ becomes 
\begin{equation}
\label{loss}
f(W_t^{'}X_t,Y_t) =  \|W_t^{'}X_t+b_t\bm{1}_t^{'}-Y_t\|_F^2.
\end{equation}
Here, $\bm{1}_t \in R^{n_t \times 1}$ is a vector with all the elements as 1.

In this case, the optimal solution to problem \eqref{MTL-1W} can be obtained by the following formula
\begin{equation}
\label{solutionW1}
W_t=(X_tX_t^{'}+\gamma FF^{'})^{-1}X_t(Y_t^{'}-\bm{1}_tb_t^{'}).
\end{equation}
And the bias vector $b_t$ can be obtained by
\begin{equation}
\label{solutionb1}
b_t = \frac{1}{n_t}Y_t \bm{1}_t - \frac{1}{n_t}W_t^{'}X_t\bm{1}_t.
\end{equation}
Note that we only need to compute $FF^{'}$ in Eq. \eqref{solutionW1} for the optimal solution $W_t$. Hence, to accelerate the proposed algorithm, we can compute $FF^{'}$ directly without computing $F$ in problem \eqref{MTL-1F}. The detailed derivation is given next.

Suppose the eigen-decomposition $WW^{'} = U\Sigma U^{'}$, $U$ is the eigenvector matrix and $\Sigma$ is the eigenvalue matrix with the order from small to large. Denote $U = [U_1,U_2,U_3]$, where $U_1 \in R^{d \times (d-c)}$, $U_2 \in R^{d \times(k-d+c)}$, $U_3 \in R^{d \times (d-k)}$. Then the optimal solution $F$ in problem \eqref{MTL-1F} is $F=[U_1,U_2]$. Note that $UU^{'} = [U_1,U_2][U_1,U_2]^{'}+U_3U_3^{'}=I$, so we have
\begin{equation}
\label{FF}
FF^{'} = [U_1,U_2][U_1,U_2]^{'} = I - U_3U_3^{'}.
\end{equation}
Due to $U_3\in R^{d\times (d-k)}$, $d-k$ is the rank of the learned $W$ which is usually much smaller than $k$. Hence, it's more efficient to utilize the formula \eqref{FF} than to calculate $FF^{'}$ directly. Based on the above derivation, an efficient optimization algorithm is obtained to solve problem \eqref{MTL-1}, and we give the detailed process in Algorithm \ref{alg1}.
\begin{algorithm}
\caption{Algorithm to solve problem \eqref{MTL-1} (KMSV) }
\label{alg1}
\begin{algorithmic}
\STATE \textbf{Input}: The training dataset $X_t \in R^{d \times n_t}$ and the label matrix $Y_t\in R^{c_t \times n_t}$ for each task $t$.
\STATE \textbf{Output}: $W \in R^{d\times c}$.
\STATE \textbf{Initialize}: $W \in R^{d\times c}$.
\REPEAT
\STATE 1. Update $FF^{'}$ by utilizing Eq. \eqref{FF}.
\STATE 2. Update $W_t$ and $b_t$ by Eq. \eqref{solutionW1} and \eqref{solutionb1} for task $t$.
\UNTIL Convergence
\end{algorithmic}
\end{algorithm}

\subsection{Optimization for Problem \eqref{MTL-2}}
\label{opt2}
Solving the problem \eqref{MTL-2} is a little difficult. If we follow the similar idea as in subsection \ref{opt1}, we can get 
\begin{equation}
\label{Kanfan2}
\sum \limits_{i=1}^k \sigma_i(W) = \min \limits_{F \in R^{d \times k}, F^{'}F=I} tr(F^{'}(WW^{'})^{\frac{1}{2}}F).
\end{equation}
It's easy to obtain the solution $F$ from problem \eqref{Kanfan2}. But when introducing this regularization into the MTL model just like the problem \eqref{MTL-1kanfan}, it's hard to optimize $W$ with the fixed $F$.

Therefore, we need to design another approach to solve problem \eqref{MTL-2}. Fortunately, we have the following equation
\begin{equation}
\label{The1}
\sum \limits_{i=1}^k\sigma_i(W)=\|W\|_*-\max \limits_{\substack{F\in R^{d\times r},F^{'}F=I\\G \in R^{c\times r},G^{'}G=I}}tr(F^{'}WG).
\end{equation}
Here, $r = \min \{d,c\}-k$. Equation \eqref{The1} can be proved by the Lagrange multiplier method with KKT condition \cite{nakayama1975generalized}. Due to the limitation of pages, the proof is not given here. Based on Eq. \eqref{The1}, problem \eqref{MTL-2} can be transformed as follows
\begin{equation}
\label{MTL-2_1}
\begin{split}
&\min \limits_{\substack{W , F, G }}\sum \limits_{t=1}^Tf(W_t^{'}X_t,Y_t)+\gamma \|W\|_*-\gamma tr(F^{'}WG) \\
&s.t. \quad F^{'}F=I, G^{'}G=I, F\in R^{d\times r}, G \in R^{c\times r}.
\end{split}
\end{equation}


\begin{algorithm}
\caption{Algorithm to solve problem \eqref{MTL-2} (KMSV-new)}
\label{alg2}
\begin{algorithmic}
\STATE \textbf{Input}: The training data matrix $X_t \in R^{d \times n_t}$ and the label matrix $Y_t \in R^{c_t \times n_t}$ for each task $t$.
\STATE \textbf{Output}: $W \in R^{d \times c}$.
\STATE \textbf{Initialize}: $W \in R^{d \times c}$.
\REPEAT
\STATE 1. Update $F$ and $G$ by solving the problem \eqref{FG}.
\STATE 2. Computing $D=\frac{1}{2}(\tilde{W}\tilde{W}^{'})^{-\frac{1}{2}}$, which is defined in the problem \eqref{MTL-2_1_Wt_reweight}.
\STATE 3. Update $W_t$ and $b_t$ by Eq. \eqref{solution2Wt} and \eqref{solutionb1} for each $t$.
\UNTIL{Converges}
\end{algorithmic}
\end{algorithm}

When $W$ is fixed, problem \eqref{MTL-2_1} becomes
\begin{equation}
\label{FG}
\max \limits_{\substack{F\in R^{d\times k}, F^{'}F=I,\\ G \in R^{c\times r}, G^{'}G=I.}}tr(F^{'}WG)\\
\end{equation}
The optimal solution $F$ and $G$ to problem \eqref{FG} are formed by the $r$ left and right singular vectors of $W$ corresponding to the $r$ largest singular values, respectively.

With $F$ and $G$ fixed, problem \eqref{MTL-2_1} can be converted into the following form
\begin{equation}
\label{MTL-2_1_Wt}
\min \limits_{W=[W_1,...,W_T]} \sum \limits_{t=1}^Tg(W_t)+\gamma \|W\|_*.
\end{equation}
Here, we define $g(W_t)=f(W_t^{'}X_t,Y_t)-\gamma tr(W_t^{'}FG_t^{'})$, and the matrix $G_t$ is the row submatrix of $G$, which corresponds to the project matrix $W_t$ for each task $t$. 

Based on the reweighted method \cite{nie2012low, nie2017multiclass}, we can solve problem \eqref{MTL-2_1_Wt} by iteratively optimizing the following problem
\begin{equation}
\label{MTL-2_1_Wt_reweight}
\min \limits_{W=[W_1,...,W_T]} \sum \limits_{t=1}^Tg(W_t)+\gamma tr(W^{'}DW).
\end{equation}
where $D = 1/2\cdot (\tilde{W}\tilde{W}^{'})^{-1/2}$, $\tilde{W}$ is the current solution of problem \eqref{MTL-2_1_Wt_reweight}. It can be proved that the proposed re-weighted based method decreases the objective value of problem \eqref{MTL-2_1_Wt} in each iteration and will converge to the optimal solution.

With the equation $tr(W^{'}DW) = \sum_t tr(W_t^{'}DW_t)$, it's easy to see that problem \eqref{MTL-2_1_Wt_reweight} is independent for different task $t$. So problem \eqref{MTL-2_1_Wt_reweight} can be divided into $T$ subproblems as 
\begin{equation}
\label{MTL-2_1_Wt_reweight2}
\min \limits_{W_t} g(W_t)+\gamma tr(W_t^{'}DW_t).
\end{equation}
Problem \eqref{MTL-2_1_Wt_reweight2} is a convex problem. Combining with the definition of $f(W_t^{'}X_t,Y_t)$ in Eq. \eqref{loss}, the optimal solution $W_t$ for different task $t$ can be obtained by
\begin{equation}
\label{solution2Wt}
W_t=(X_tX_t^{'}+\gamma D)^{-1}(X_t(Y_t^{'}-\bm{1}b_t^{'})+\frac{1}{2}\gamma FG_t^{'}).
\end{equation}
The bias vector $b_t$ can also be calculated by formula \eqref{solutionb1}. The whole process to solve problem \eqref{MTL-2} is summarized in Algorithm \ref{alg2}.

\begin{figure}[ht]
  \centering
  \subfigure[The value of nMSE]{
    \label{Fig.1a}
     \includegraphics[scale=0.305]{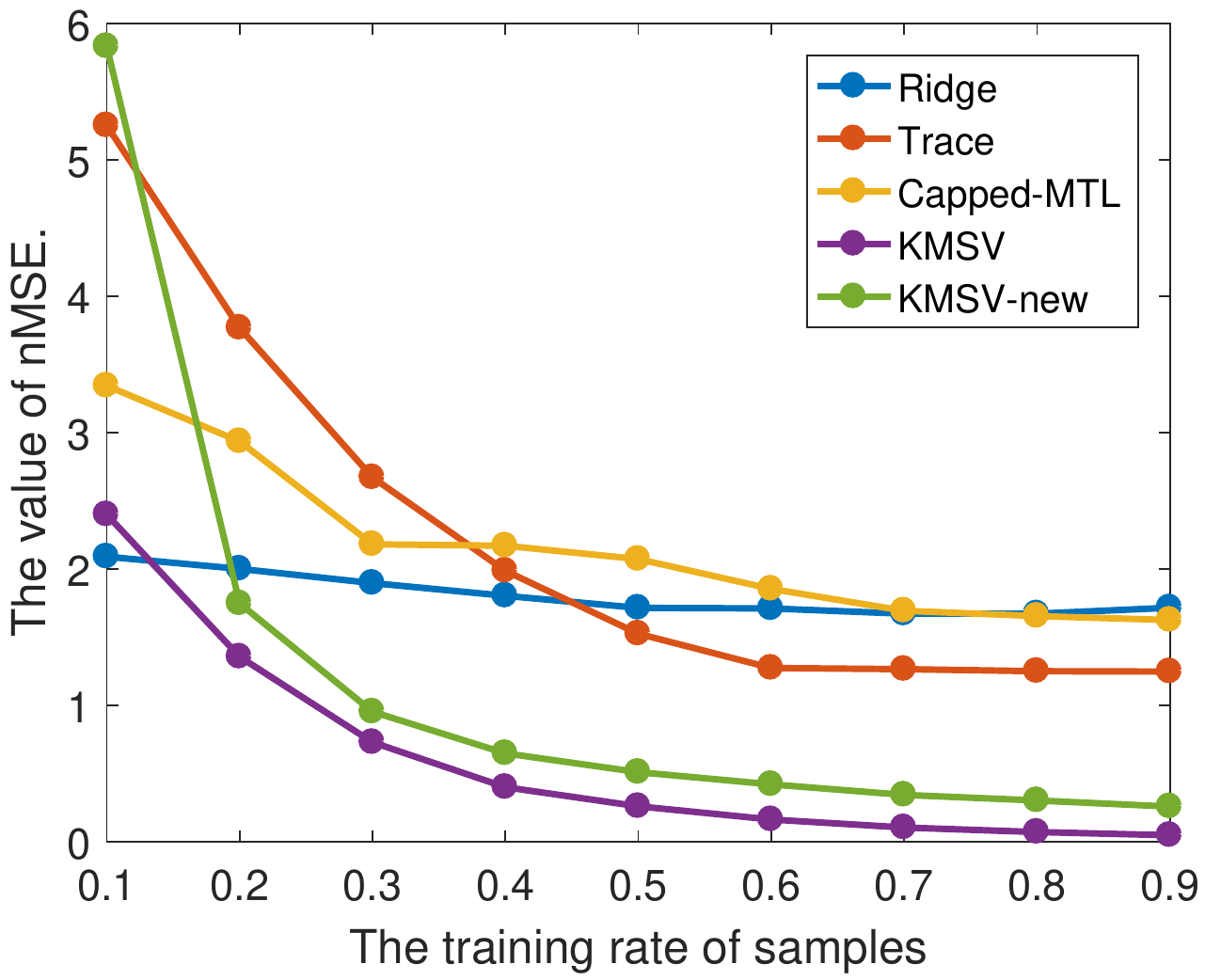}}
  \subfigure[The value of $E.W.$]{
    \label{Fig.1b}
      \includegraphics[scale=0.305]{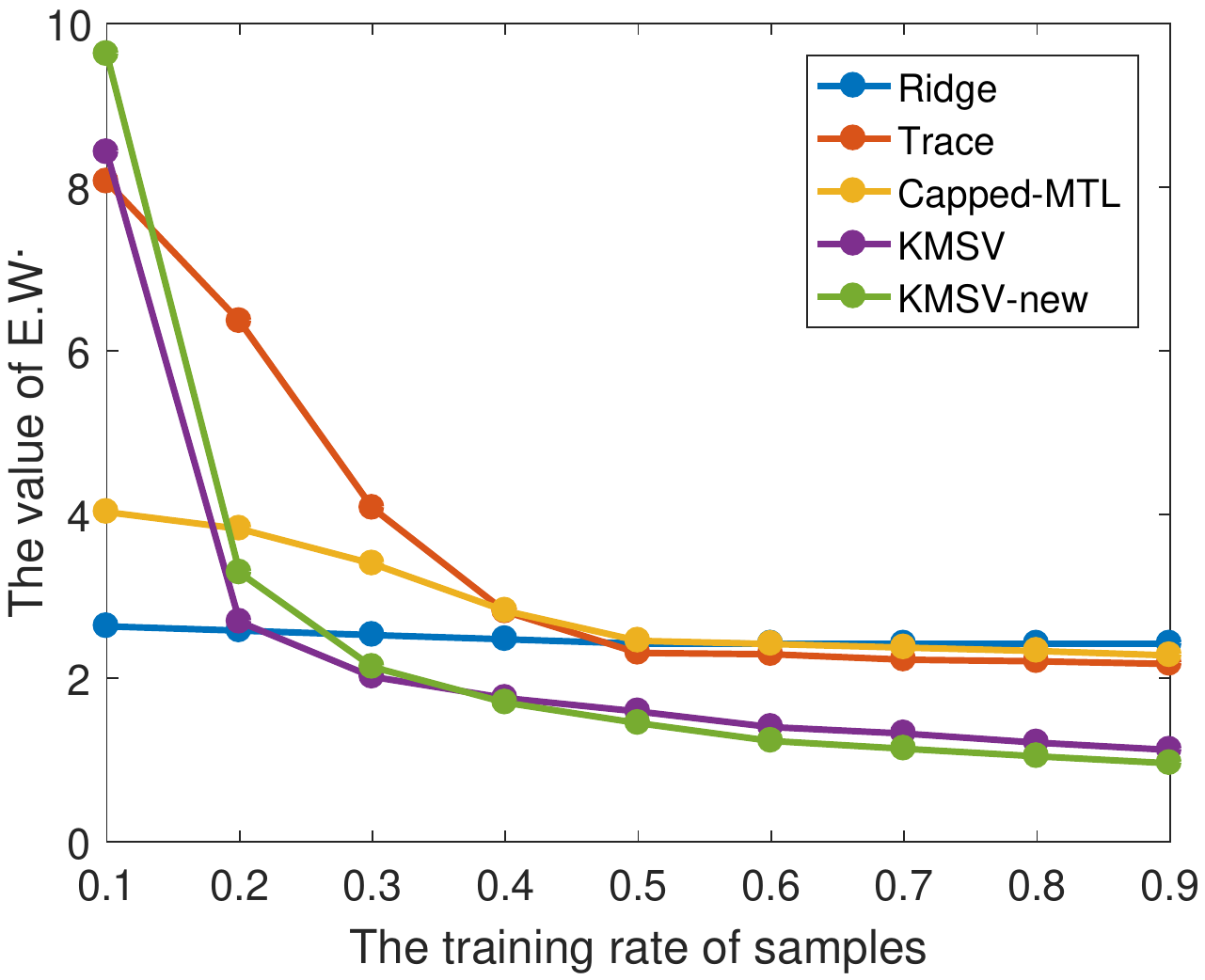}}
  \caption{The change curves of averaged nMSE and E.W. by the comparison methods under the different training rates. } 
  \label{fig1}
\end{figure}

\begin{figure}[h]
  \centering
  \subfigure[The singular values of $W$]{
    \label{Fig.2a}
     \includegraphics[scale=0.302]{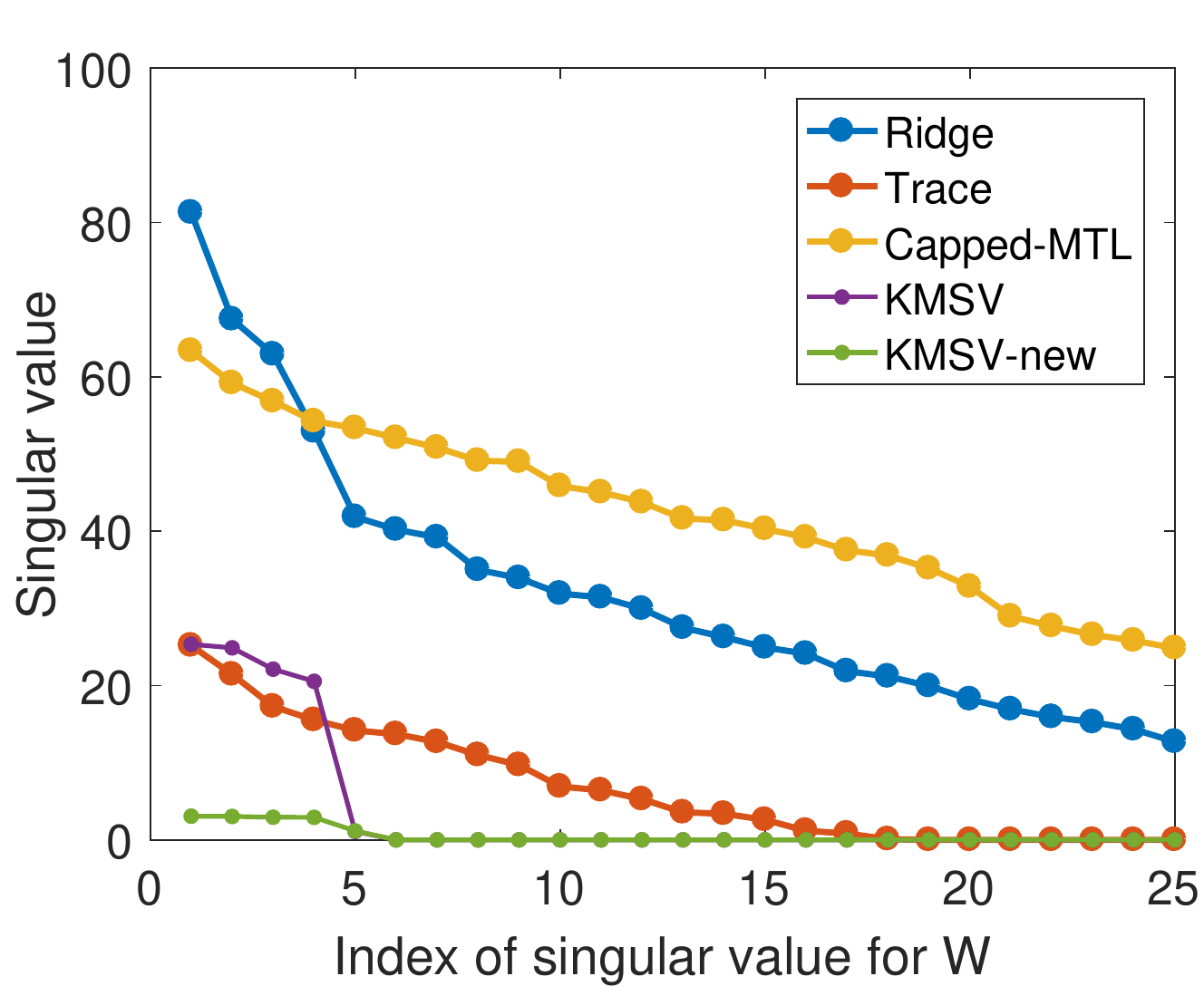}}
  \subfigure[The convergence curves]{
    \label{Fig.2b}
      \includegraphics[scale=0.302]{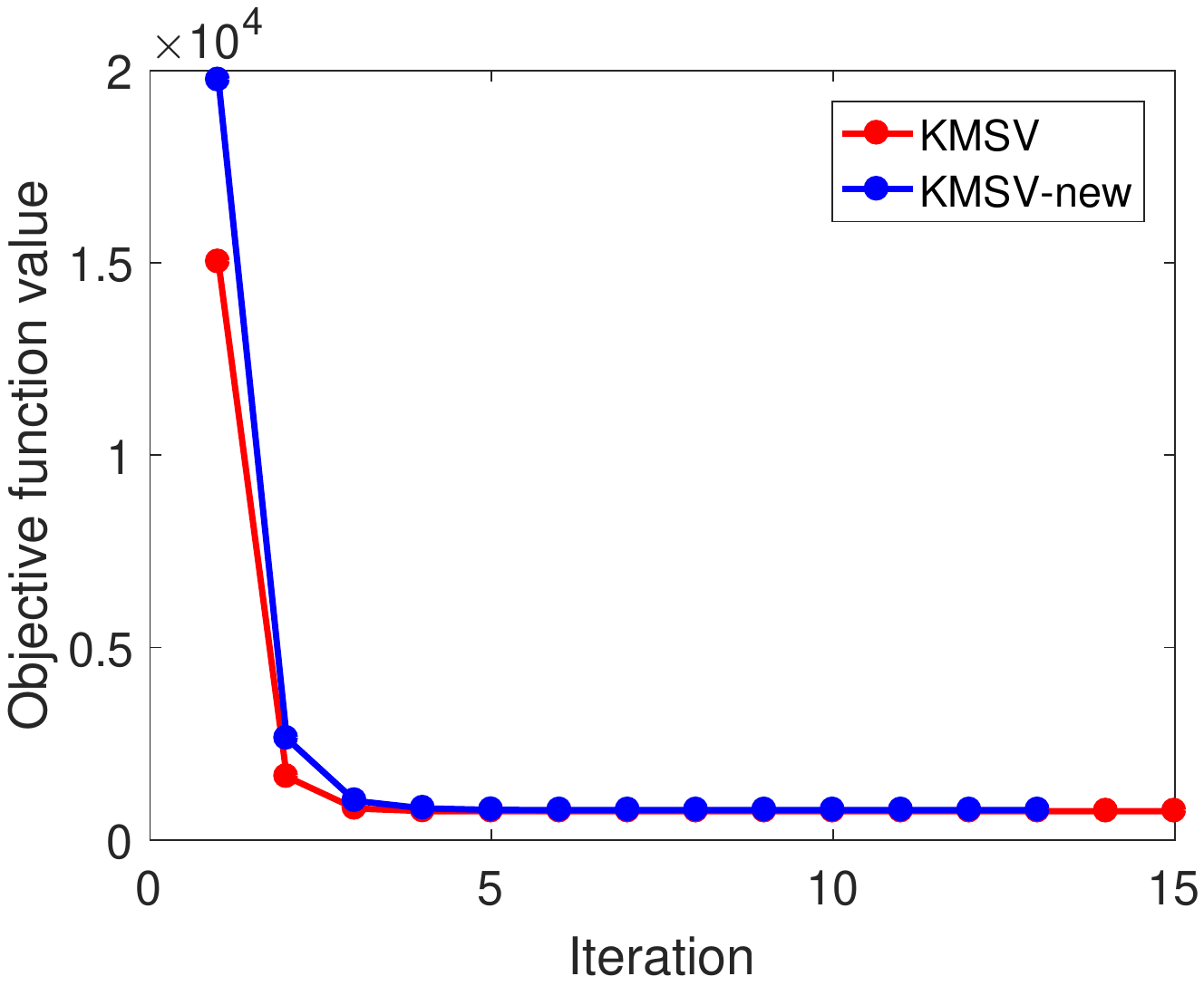}}
  \caption{(a). The distribution of singular values for $W$ obtained by five algorithms under the training rate 50\%. (b). The convergence curve of our models under the training ratio of 50\%.} 
  \label{fig2}
\end{figure}

\begin{table*}[t]
\centering
  \begin{threeparttable} 
\caption{\normalsize{The comparison results on SCHOOL dataset based on the evaluation metric nMSE with standard deviation.}} 
\label{table}
\small
\renewcommand\tabcolsep{2.0pt}
\renewcommand\arraystretch{1.25}
\begin{tabular}{|c|c|c|cccccc}
\hline
\multirow{2}{*}{Ratio} & \multicolumn{2}{c|}{Single Task Method}                           & \multicolumn{6}{c|}{Low-Rank Based MTL Method}                                                                                                                                                    \\ \cline{2-9} 
                       & Ridge    & Lasso  & \multicolumn{1}{c|}{Trace}  & \multicolumn{1}{c|}{Capped-MTL} & \multicolumn{1}{c|}{NN-MTL} & \multicolumn{1}{c|}{CMTL}   & \multicolumn{1}{c|}{KMSV} & \multicolumn{1}{c|}{KMSV-new}  \\ \hline \hline
10\%                   & 34.1317(6.1479) & 5.4637(1.0372) & \multicolumn{1}{c|}{3.2138(0.6051)} & \multicolumn{1}{c|}{1.3771(0.0445)} & \multicolumn{1}{c|}{2.4557(0.0226)} & \multicolumn{1}{c|}{1.5148(0.0598)} & \multicolumn{1}{c|}{\underline{1.3082(0.0338)}} & \multicolumn{1}{c|}{\textbf{1.1265(0.0292)}} \\ \hline
20\%                   & 21.3752(2.9872) & 4.9860(0.6474)  & \multicolumn{1}{c|}{1.1756(0.2446)} & \multicolumn{1}{c|}{1.0844(0.0257)} & \multicolumn{1}{c|}{2.4207(0.0088)} & \multicolumn{1}{c|}{1.1567(0.0377)} & \multicolumn{1}{c|}{\underline{1.0480(0.0225)}}  & \multicolumn{1}{c|}{\textbf{0.9699(0.0194)}} \\ \hline
30\%                   & 16.2509(1.7733) & 4.2286(0.2885) & \multicolumn{1}{c|}{1.1551(0.0427)} & \multicolumn{1}{c|}{0.9978(0.0385)} & \multicolumn{1}{c|}{1.7045(0.0916)} & \multicolumn{1}{c|}{1.0379(0.0436)} & \multicolumn{1}{c|}{\underline{0.9695(0.0300)}} & \multicolumn{1}{c|}{\textbf{0.9099(0.0100)}} \\ \hline
\end{tabular}
\end{threeparttable}  
\end{table*}

\section{Experiment}
\label{experi}
In this section, we will verify the proposed MTL models denoted as KMSV and KMSV-new on two benchmark datasets. The metric nMSE \cite{nguyen2009local} is adopted as the evaluation index, of which the smaller value means the better performance. 
The parameter $\gamma$ is set as $10^2$ for KMSV, $10^4$ for KMSV-new.

\textbf{Synthetic Data.} Based on the method referred in \cite{nie2018calibrated}, we build a synthetic dataset $\{X_t\}_t^T$ consisting of $T=25$ regression tasks. These tasks are all generated by a 100-dimensional Gaussian distribution randomly, and we set the number of samples per task to 400. The projection matrix $W^* \in R^{d \times T}$ is generated with the rank of 5. Hence, we can set $k=20$ in our models. The truth label $Y_t$ is obtained by $W^*$ and $X_t$. The normal Gaussian noise is also added to $Y_t$. Due to the given matrix $W^*$, another criteria $E.W. = \|W-W^*\|_F^2/T$ similar to nMSE is introduced to evaluate the algorithms. 

Comparing with Ridge regression (Ridge) \cite{hoerl1970ridge}, Trace Norm Minimization (Trace) \cite{ji2009accelerated} and Capped-MTL \cite{han2016multi}, Figure \ref{fig1} presents the change curves of nMSE and E.W. under the different training rates. We can see that KMSV and KMSV-new achieve the smaller values of nMSE and E.W. than other methods. It means our algorithms KMSV and KMSV-new are better at capturing the low-rank structure, which can be further demonstrated by the distribution of singular values in the obtained $W$ shown in Figure \ref{Fig.2a}. Besides, Figure \ref{Fig.2b} presents the convergence of KMSV and KMSV-new, which illustrates that our models are efficient to deal with this synthetic dataset.

\textbf{SCHOOL Dataset.} This dataset \cite{chen2011integrating} contains 139 regression tasks, each of which has the same 28 features. We randomly select 10\%, 20\% and 30\% of the samples from each task to form the training set and the rest is the test set. Besides, in the process of training, to further verify that one single task can benefit from the co-training, we randomly select 30\% tasks and add the white noise to their training labels. Here, the white noise is drawn from the Gaussian distribution $\mathcal{N}(1,2)$. In this section, we compare our models with single-task models: Ridge, Lasso \cite{tibshirani1996regression} and low-rank based MTL methods: Trace, Capped-MTL, NN-MTL \cite{chen2009convex} and CMTL \cite{nie2018calibrated}. The parameters in all comparison methods are tuned to the best through the corresponding references and the hyperparameter $k$ of our models is set to 10.

The presented algorithms are all conducted on SCHOOL dataset ten times and Table \ref{table} gives the mean value of nMSE with its standard error. From Table \ref{table}, we notice that our models are better than the low-rank based MTL methods. Besides, KMSV-new and KMSV achieve the best and second-best results among all comparison methods, respectively. Hence, it can be concluded that multi-task learning can improve the learning performance of single task effectively. Furthermore, due to the novel regularization terms, our methods have the superior learning ability to other MTL methods in practical circumstance.

\section{Conclusion}
\label{concl}
In this paper, we propose two novel multi-task learning models KMSV and KMSV-new based on the designed regularization terms, and apply them in regression problem. The new proposed regularization terms are the better approximation of the rank minimization problem, which makes our methods capture the low-rank structure shared across tasks efficiently, and outperform other classical MTL methods. An efficient algorithm based on the iterative re-weighted method is proposed to optimize our models. Experimental results on synthetic and real-world datasets demonstrate the superiority of our methods. For our models, we don't have to adjust the parameter $\gamma$ specifically. But the hyperparameter $k$ still needs to be tuned by mankind. So we need to design an efficient strategy to determine the hyperparameter $k$ in future work.

\bibliographystyle{ACM-Reference-Format}
\balance
\bibliography{reference}

\end{document}